%% file: main.tex
\documentclass[sigconf,review=False]{acmart}
\AtBeginDocument{
  \providecommand\BibTeX{{
    \normalfont B\kern-0.5em{\scshape i\kern-0.25em b}\kern-0.8em\TeX}}}

\input{sections/00-preamble}
\input{sections/0a-macros}

\begin{document}

\input{sections/0b-title-and-authors}
\input{sections/1-abstract}

\maketitle

\input{sections/2-intro}
\input{sections/3-claimbuster}
\input{sections/4-adversarial}
\input{sections/5-results-discussion}
\input{sections/6-related-works}
\input{sections/7-conclusion}

\bibliographystyle{ACM-Reference-Format}
\bibliography{biblio}

\input{sections/9-reproducibility}

\end{document}

%% file: sections/00-preamble.tex
\usepackage[ruled,vlined]{algorithm2e}
\usepackage[multiple,stable,para]{footmisc}
\usepackage[normalem]{ulem}
\usepackage{cancel}
\usepackage{relsize}
\usepackage{multirow}
\usepackage{tikz}
\usepackage{graphicx}
\usepackage{colortbl}
\usepackage{amsmath}
\usepackage{mathtools}
\usepackage{siunitx}
\usepackage{changepage}
\usepackage{wasysym}
\usepackage[skip=2pt]{caption}
\usepackage{cleveref}
\crefformat{footnote}{#2\footnotemark[#1]#3}


\renewcommand\footnotetextcopyrightpermission[1]{}

%% file: sections/0a-macros.tex
\newcommand{\ExternalLink}{
    \tikz[x=1.2ex, y=1.2ex, baseline=-0.05ex]{
        \begin{scope}[x=1ex, y=1ex]
            \clip (-0.1,-0.1) 
                --++ (-0, 1.2) 
                --++ (0.6, 0) 
                --++ (0, -0.6) 
                --++ (0.6, 0) 
                --++ (0, -1);
            \path[draw, 
                line width = 0.5, 
                rounded corners=0.5] 
                (0,0) rectangle (1,1);
        \end{scope}
        \path[draw, line width = 0.5] (0.5, 0.5) 
            -- (1, 1);
        \path[draw, line width = 0.5] (0.6, 1) 
            -- (1, 1) -- (1, 0.6);
        }
    }

\definecolor{mygreen}{RGB}{34,139,34}

%% file: sections/0b-title-and-authors.tex
\title[Gradient-Based Adversarial Training on Transformer Networks for Detecting Check-Worthy Factual Claims]{Gradient-Based Adversarial Training on Transformer Networks for Detecting Check-Worthy Factual Claims}

\author{Kevin Meng}
\authornote{The participation of Meng in this work started in a 2018 summer camp program hosted by UT-Arlington and continued afterwards in the author's capacity as an affiliate of the university. Meng is expected to start his undergraduate study at MIT in Fall 2020.}
\authornote{Meng and Jimenez made equal contribution to this work and thus share co-first authorship.\\}
\affiliation{
 \institution{Plano West Senior High School}
 \city{Plano}
 \state{TX}
 \country{USA}}
\email{kevinmeng@acm.org}

\author{Damian Jimenez}
\authornotemark[2]
\affiliation{
 \institution{The University of Texas at Arlington}
 \city{Arlington}
 \state{TX}
 \country{USA}}
\email{damian.jimenez@mavs.uta.edu}

\author{Fatma Arslan}
\affiliation{
 \institution{The University of Texas at Arlington}
 \city{Arlington}
 \state{TX}
 \country{USA}}
\email{fatma.dogan@mavs.uta.edu}

\author{Jacob Daniel Devasier}
\affiliation{
 \institution{The University of Texas at Arlington}
 \city{Arlington}
 \state{TX}
 \country{USA}}
\email{jacob.devasier@mavs.uta.edu}

\author{Daniel Obembe}
\affiliation{
 \institution{The University of Texas at Arlington}
 \city{Arlington}
 \state{TX}
 \country{USA}}
\email{daniel.obembe@mavs.uta.edu}

\author{Chengkai Li}
\affiliation{
 \institution{The University of Texas at Arlington}
 \city{Arlington}
 \state{TX}
 \country{USA}}
\email{cli@uta.edu}

%% file: sections/1-abstract.tex
\begin{abstract}

We present a study on the efficacy of adversarial training on transformer neural network models, with respect to the task of detecting check-worthy claims. In this work, we introduce the first adversarially-regularized, transformer-based claim spotter model that achieves state-of-the-art results on multiple challenging benchmarks. We obtain a $4.70$ point F1-score improvement over current state-of-the-art models on the ClaimBuster Dataset and CLEF2019 Dataset, respectively. In the process, we propose a method to apply adversarial training to transformer models, which has the potential to be generalized to many similar text classification tasks. Along with our results, we are releasing our codebase and manually labeled datasets. We also showcase our models' real world usage via a live public API.~\cref{claimbuster-url}

\end{abstract}

%% file: sections/2-intro.tex
\begin{figure*}[!htb]
  \centering
  \includegraphics[keepaspectratio, width=0.85\textwidth]{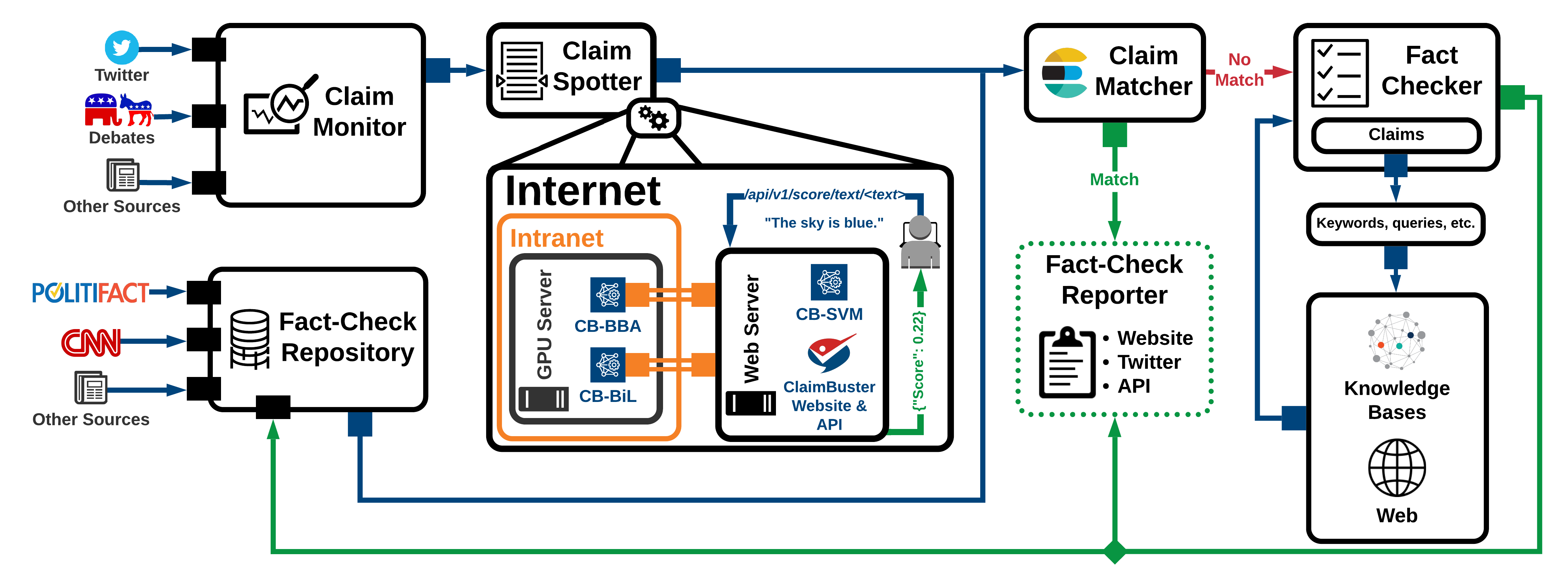}
  \caption{A Diagram of Our Current Fact-Checking Framework}
  \label{fig:claimbuster-framework}
\end{figure*}

\section{Introduction} \label{sec:intro}

The creation and propagation of misinformation has become an increasingly important issue for our society to tackle. Today, many falsehoods are spread via mediums that allow quick dissemination of information, including social media, news outlets, and televised programs. The distribution of objectively incorrect information can negatively impact the operation of our society in many spheres. Especially in the realm of political discourse, misinformation can shake public confidence in government institutions,~\footnote{\url{https://pewrsr.ch/2HoH0au}} erroneously inform political judgements~\cite{NBERw23089}, and reinforce confidence in wrong information~\cite{Chan_2017}.

In recent years, the number of fact-checking outlets has grown from 44 in 2014 to 226 in 2019~\footnote{\url{https://reporterslab.org/tag/fact-checking-database/}} as part of a global movement to suppress misinformation. These outlets, including PolitiFact,~\footnote{\label{politifact-url}https://www.politifact.com/} Snopes,~\footnote{\label{snopes-url}https://www.snopes.com/} and FactCheck.org,~\footnote{\label{factcheckorg-url}\url{https://www.factcheck.org/}} hire human fact-checkers to perform \textit{claim-checking}, a process in which they vet factual claims by reviewing relevant source documents and interviewing subject experts. In addition to outlets that directly fact-check claims, there exist many projects that use computing to aid fact-checkers in disseminating information to the general public, including Schema.org's ClaimReview~\footnote{\label{claimreview-url}\url{https://schema.org/ClaimReview}} which organizes fact-checks into a unified database; FactStream~\footnote{\label{factstream-url}\url{https://www.factstream.co/}} which compiles fact-checks into a smartphone application; and Fatima,~\footnote{\label{fatima-url}\url{https://fatima.aosfatos.org/}} a bot built by Aos Fatos, a Brazilian fact-checking organization, that scans Twitter for tweets containing already-debunked misinformation and refers readers to relevant fact-checks. These organizations and projects play a central role in fighting misinformation, as fact-checks are effective not only for debunking false claims but also deterring speakers from making false claims in the future~\cite{nyhan2015estimating}.

However, due to the intense time commitment demanded by fact-checking, combined with the rapid rate at which new content surfaces via modern media channels, many problematic claims go unnoticed and unchecked~\cite{pennycook2019fighting}. These challenges present an opportunity for \textit{automated} fact-checking tools to help fact-checkers perform their duties. There are several prominent fact-checking projects that are currently testing automated systems, including FactChecker~\footnote{\label{factchecker-url}\url{https://idir.uta.edu/claimbuster-dev/factchecker/}} which queries knowledge bases, cross-references known fact-checks, and provides custom ranked Google search results; ClaimPortal ~\footnote{\label{claimportal-url}\url{https://idir.uta.edu/claimportal/}} which uses ClaimBuster~\cite{Hassan2017claimbusterfull, jimenez2018claimspot, Hassan2017KDDclaimspot, hassan2015detecting} to select tweets that are worth fact-checking, as well as various algorithms to retrieve relevant articles and pre-existing fact-checks~\cite{majithia2019claimportal};  Squash~\footnote{\label{squash-url}\url{https://bit.ly/31YTfnJ}} which fact-checks live debates by converting speech to text and querying a database of pre-existing fact-checks; Fakta~\footnote{\label{fatka-url}\url{https://fakta.app/}} which checks claims against reliable web sources~\cite{nadeem2019fakta}; and FullFact~\footnote{\label{fullfact-url}\url{https://fullfact.org/automated}} which is developing systems to cluster groups of similar claims together.

Claim-spotting is a process that precedes claim-checking where check-worthy claims are spotted from large streams of information available from various sources (e.g., newscasts, news websites, Twitter, Facebook). Claim-spotting is an area that is highly suitable for machine learning algorithms to tackle. The work presented here focuses on the claim-spotting component of ClaimBuster,~\footnote{\label{claimbuster-url}\url{https://idir.uta.edu/claimbuster/api/docs/}} which scores claims based on their check-worthiness. This is paramount to ensuring that 1) check-worthy factual claims are not missed by fact-checkers and 2) unimportant or non-factual claims do not congest fact-checkers' intellectual bandwidth. To this day, ClaimBuster's API is regularly in use not only by internal projects such as ClaimPortal but also external collaborators such as the Duke Reporters' Lab. The closest projects to ClaimBuster in this space are QCRI's ClaimRank~\footnote{\label{claimrank-url}\url{https://claimrank.qcri.org/}} project, and a component in FullFact's proposed automated fact-checking system which they gave the name Hawk in their whitepaper.~\footnote{\label{fullfactwhitepaper-url}\url{https://bit.ly/31YTsY3}} QCRI's ClaimRank is very similar to ClaimBuster in that it ranks claims by assigning them a check-worthiness score from $0$ to $1$. As for Hawk and FullFact's system, not many details have been released.

Currently, no existing claim-spotter~\cite{Hassan2017KDDclaimspot, hassan2015detecting, jimenez2018claimspot, Copenhagen-team, TheEarthIsFlat-team} has attempted to apply transformers \cite{vaswani2017attention} to the claim-spotting task. The transformer is a new deep learning architecture that has recently allowed for rapid progress and development in the natural language processing field. Particularly, Bidirectional Encoding Representations from Transformers (BERT)~\cite{devlin2018bert} has achieved state-of-the-art performance on many challenging language understanding and classification benchmarks. We surmise that BERT's architecture is suited for our claim-spotting task. However, BERT models have upwards of 300 million trainable parameters, making them highly susceptible to overfitting \cite{caruana2001overfitting}, especially on limited amounts of training data. To address this, we propose to incorporate \textit{adversarial training} into a BERT-based model as a regularization technique. Gradient-based adversarial training \cite{goodfellow2014explaining, miyato2016adversarial, miyato2018virtual} is a procedure that trains classifiers to be resistant to small, approximately worst-case perturbations to its inputs. It was first applied to computer vision tasks in \cite{goodfellow2014explaining} and later brought to the NLP domain in Long Short-Term Memory Networks~\cite{hochreiter1997long} by Goodfellow et al.~\cite{miyato2016adversarial, miyato2018virtual}. No prior work has attempted to incorporate this type of adversarial training into transformer networks. We are the first to propose this technique, which is also potentially applicable in many other NLP-related tasks.

Motivated by the above, we introduce the first adversarially-regularized, transformer-based claim-spotting model that achieves state-of-the-art results on challenging claim-spotting benchmarks. Our contributions are summarized as follows:
\begin{itemize}
    \item We are the first to apply gradient-based adversarial training to transformer networks. 
    \item We present the first transformer-based neural network architecture for claim-spotting.
    \item Our models are the first claim-spotters to be regularized by gradient-based adversarial training.
    \item Our models achieve state-of-the-art performance by a substantial margin on challenging claim-spotting benchmarks.
    \item We release a public codebase, dataset, and API for both reproducibility and further development (Section \ref{sec:repro}).
\end{itemize}

%% file: sections/3-claimbuster.tex
\section{ClaimBuster Overview} \label{sec:claimbuster-overview}
In this section we present a brief history and overview on the ClaimBuster project. We cover its inception and impact in the community, as well as the current status of our fact-checking framework.

\subsection{ClaimBuster's History and Current Status} \label{sec:claimbuster-history}
ClaimBuster's foundation was first established in \cite{hassan2015detecting}, where Hassan \textit{et. al.} first presented results on different machine learning models trained on an early version of the dataset we are using currently. This work later evolved into what is currently known as ClaimBuster and was presented in \cite{Hassan2017KDDclaimspot,Hassan2017claimbusterfull}. Since then, ClaimBuster has partnered with the Duke Reporters' Lab (DRL)~\footnote{\label{duke-reporters-lab-url} \url{https://reporterslab.org/}} and collaborated with them through ClaimBuster's API. During this time ClaimBuster's API has been called over $\mathbf{47,000,000}$ times by internal projects and over $\mathbf{456,000}$ times by the DRL. ClaimPortal is the internal project that has made the most use of the ClaimBuster API. It scores tweets and provides relevant fact-checks for tweets using the claim-matching component seen in Figure~\ref{fig:claimbuster-framework}. Through this project we have seen that we can successfully apply ClaimBuster to different domains, such as Twitter. As for the DRL, they generate and send out a daily e-mail to fact-checkers with the latest top claims that were identified by ClaimBuster from television and social media. Through our collaboration with the DRL we have been able to contribute to which claims are fact-checked by major news outlets.~\footnote{\label{washingtonpost-claimbuster-url} \url{https://bit.ly/2vs8Fol}} The accessibility of our work has allowed it, in general, to have a widespread impact in the fact-checking community.

Since the development of the original SVM model, we have been exploring deep learning~\cite{jimenez2018claimspot} approaches to improve our claim-spotting model. Recently, this culminated with us employing the BERT architecture due to BERT and its derivative models' proven track record in performing well on NLP related tasks such as SQuAD and GLUE.~\footnote{\label{squad-url}\url{https://rajpurkar.github.io/SQuAD-explorer/}}~\footnote{\label{glue-url} \url{https://gluebenchmark.com/leaderboard}} Since then, we have also re-evaluated our approach to the classes used within our datasets, how our dataset is generated, and refined our overall process when it comes to evaluating models. This turnaround in our approach to our dataset has come after a lengthy evaluation of our extraction criteria (i.e., what we consider high-quality labels), and the ratio of check-worthy to non-check-worthy sentences in the dataset. Through these evaluations, we are confident we have obtained a better quality dataset than that used in previous works. The work presented here will also begin a thorough test period with our collaborators at the DRL.

\subsection{Fact-Checking Framework} \label{sec:claimbuster-framework}

Figure \ref{fig:claimbuster-framework} showcases the current status of our fact-checking framework. We monitor claims from various sources (e.g., Twitter, political debates, etc.), and we are even able to process live television closed-caption feeds for important events such as presidential debates. ClaimSpotter then handles scoring all of the claims that our claim monitoring system captures. ClaimSpotter is accessible to the public via an API, which only requires a free API key~\footnote{\label{apikey-signup-url} \url{https://idir.uta.edu/claimbuster/api/request/key/}} to access. We are deploying the deep learning models for the public and other researchers to test and verify the models presented in this paper. Each deep-learning model is running off of a dedicated Nvidia GTX 1080Ti. All resources are running on the same network, so there is no significant overhead added by a server to server communication.

In addition, we also have a repository of fact-checked claims which we use in conjunction with ElasticSearch~\footnote{\label{elasticsearch-url} \url{https://www.elastic.co/}} in our claim-matcher component to verify the veracity of any claims that have been previously fact-checked by professional fact-checkers. If no previous fact-checks are found then we can send these claims to our fact-checking component, which is still being developed. Currently, our approach is to convert claims to questions~\cite{heilman-phd} in order to query knowledge bases (e.g., Wolfram, Google, etc.) using natural language to see if they can generate a clear verdict. This approach is useful for general knowledge type claims, but nuanced claims requiring domain-specific knowledge are still very challenging to handle. Finally, we also provide re-ranked Google search results which are sorted based on the content of the pages the initial search query returns. The analysis is based on the Jaccard similarity of the context surrounding the text in each page that matched the initial query. Finally, we regularly publish presidential debate check-worthiness scores during election cycles on our website,~\footnote{\label{claimbuster-debates-url} \url{https://idir.uta.edu/claimbuster/debates}} and we also post high-scoring claims on our project's Twitter account.~\footnote{\label{claimbuster-twitter-url} \url{https://twitter.com/ClaimBusterTM}}

%% file: sections/4-adversarial.tex
\begin{figure*}[!ht]
  \centering
  \includegraphics[keepaspectratio, width=0.85\textwidth]{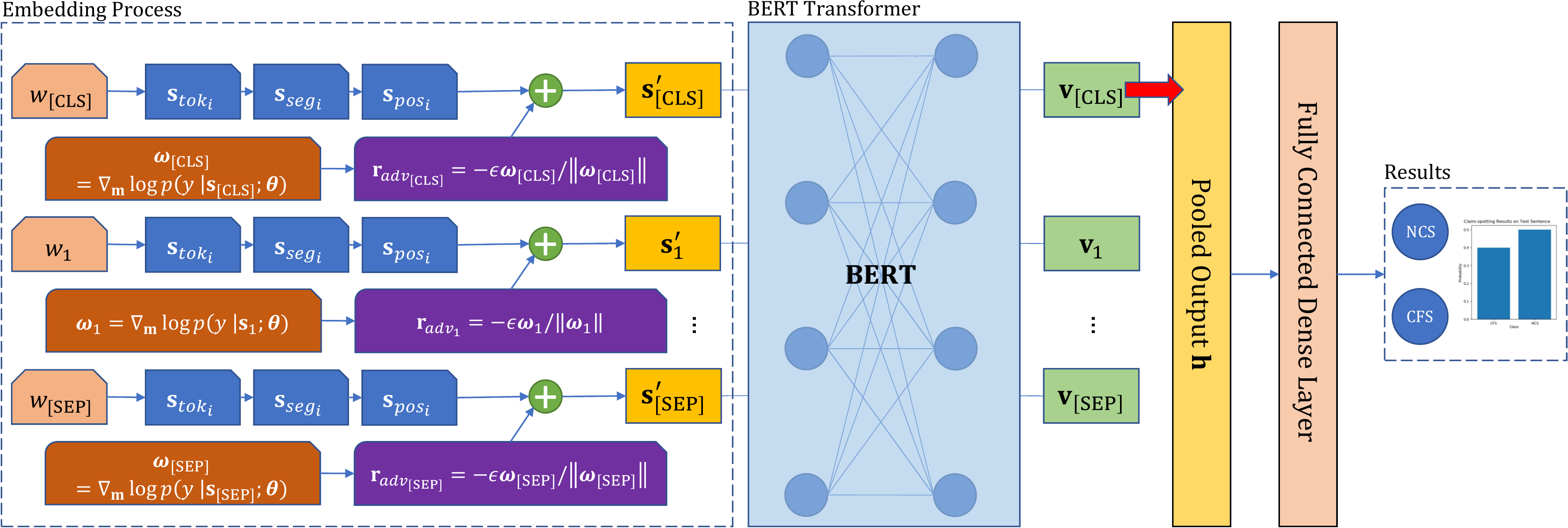}
  \caption{Our Custom Adversarially Perturbed Claim Spotting Architecture}
  \label{fig:BERTadv}
\end{figure*}

\section{BERT Claim Spotting Model} 
\label{sec:advtrain}

In this section, we present our approach to integrating adversarial training into the BERT architecture for the claim spotting task. To the best of our knowledge, our work is the first to apply gradient-sign adversarial training~\cite{goodfellow2014explaining} to transformer networks. 

\subsection{Preliminaries} \label{subsec:prelim}

\subsubsection{Task Definition} \label{subsubsec:taskdef}

Detecting check-worthy factual claims has been studied as a binary/ternary classification task and a ranking task, as explained below. In this paper, we evaluate the performance of our models on the binary and ranking task definitions.

\medskip \noindent \textbf{Binary Classification Task}: In this work a sentence $\mathbf{w}$ is classified as one of two classes, which deviates from the previous definition used in~\cite{Hassan2017claimbusterfull, jimenez2018claimspot, Hassan2017KDDclaimspot, hassan2015detecting}. 

\begin{itemize}
    \item \textbf{Non-Check-Worthy Sentence (NCS):} This class includes sentences that contain subjective or opinion-centered statements, questions, and trivial factual claims (e.g., The sky is blue).
    
    \item \textbf{Check-Worthy Factual Sentence (CFS):} This class contains claims that are both factual and salient to the general public. They may touch on statistics or historical information, among other things.
\end{itemize}

\medskip \noindent \textbf{Ranking Task}: To capture the importance of prioritizing the most check-worthy claims, a check-worthiness score~\cite{Hassan2017KDDclaimspot} is defined for each sentence $\mathbf{w}$:
\begin{equation} \label{eq:checkworthy}
    CWS = p(y = \text{CFS} \mid \mathbf{w})
\end{equation}

\noindent The $CWS$ score defines a classification model's predicted probability that a given claim is in the CFS class.

\subsubsection{BERT Language Model} \label{subsec:bert}

Bidirectional Encoder Representations from Transformers (BERT) \cite{devlin2018bert} is a transformer-based language modeling architecture that has recently achieved state-of-the-art performance on many language modeling and text classification tasks, including the Stanford Question Answering Dataset (SQuAD)~\cite{Rajpurkar_2016} and General Language Understanding Evaluation (GLUE)~\cite{wang2018glue}. We review BERT's relevant features below.

\medskip \noindent \textbf{Input/Output Representations:} Consider an arbitrary training sentence $\mathbf{w}$ with ground-truth label $y$. $\mathbf{w}$ is first tokenized using the WordPiece Tokenizer \cite{wu2016googles}.
Next, a [CLS] token is prepended to $\mathbf{w}$ to indicate the start of the sentence, a [SEP] token is appended to $\mathbf{w}$ to indicate the end of a sentence, and $\mathbf{w}$ is padded to a length of $T = 200$ using whitespace.
Each resulting token is then converted into its corresponding index in the WordPiece vocabulary list. This input vector, denoted $\mathbf{x} \in \mathbb{R}^T$, is passed to the embedding layers.

\smallskip \noindent \textbf{Three-Part Embeddings:} $\mathbf{x}$ is first transformed from a sparse bag-of-words form to a dense vector representation \cite{mikolov2013distributed} through an embedding lookup matrix $\mathbf{V} \in \mathbb{R}^{Q \times H}$, where $Q$ is the size of the WordPiece vocabulary list and $H$ is the embedding dimensionality. The series of operations that applies $\mathbf{V}$ to $\mathbf{x}$ is called the \textit{token} embedding layer, and its output is given as $\mathbf{s}_{tok} = \mathbf{V}_{x_t}, \, \forall x_t \in \mathbf{x}$, where $\mathbf{s}_{tok} \in \mathbb{R}^{T \times H}$. Additionally, BERT utilizes an \textit{segment} embedding layer that signifies which parts of the input contain the input sentence, as the end of $\mathbf{x}$ may be padded with empty space. The output of this layer is denoted by $\mathbf{s}_{seg} \in \mathbb{R}^{T \times H}$. Finally, since vanilla transformers analyze all tokens in parallel and therefore cannot account for the sequential ordering of words, BERT introduces a randomly-initialized real-valued signal via the \textit{positional} embedding layer to encode the relative order of words. The output of this layer is denoted $\mathbf{s}_{pos} \in \mathbb{R}^{T \times H}$. The final input, denoted $\mathbf{s}$, is the element-wise addition of the three separate embedding layers' outputs: $\mathbf{s} = \mathbf{s}_{tok} + \mathbf{s}_{seg} + \mathbf{s}_{pos}$.  We denote the vector representation of the $t$th token in $\mathbf{x}$ to be $\mathbf{s}_t \in \mathbb{R}^H$.

\smallskip \noindent \textbf{Transformer Encoder:} Using multiple stacked layers of attention heads~\cite{vaswani2017attention}, the BERT module encodes each input embedding $\mathbf{s}_t$ into a hidden vector $\mathbf{v}_t \in \mathbb{R}^H$, which is a hidden representation that incorporates context from surrounding words bidirectionally, as opposed to unidirectional encoders used in OpenAI GPT \cite{radford2018gpt, radford2019language}.

\smallskip \noindent \textbf{Pooling Layer:} The pooling layer generates a representation for the entire sentence by applying a dense layer on top of the [CLS] token's hidden representation, resulting in $\mathbf{h} \in \mathbb{R}^H$. This sentence-level encoding vector can be used to perform many downstream tasks including claim-spotting.

\subsection{Model Architecture} \label{sec:architecture}

In this section, we outline how BERT is integrated with adversarial perturbations to create a claim-spotting model. The resultant model is end-to-end differentiable and trainable by gradient descent~\cite{kingma2014adam}. We refer the reader to Figure \ref{fig:BERTadv} for illustrations on each of the architectural components.

\subsubsection{Embedding Process}

All three embeddings from the BERT architecture are carried over: token $\mathbf{s}_{pos}$, segment $\mathbf{s}_{seg}$, and positional $\mathbf{s}_{pos}$. Each embedding layer still performs its original function, transforming a given word $\mathbf{x}$ into the embedding representation $\mathbf{s}$. The key difference in our architecture is the implantation of an addition gate through which adversarial perturbations $\mathbf{r}_{adv}$ are injected into $\mathbf{s}$ to create the perturbed embedding $\mathbf{s}^\prime$. 

\subsubsection{BERT Transformer}

Our work harnesses the power of the BERT architecture which supports transfer learning \cite{tan2018transferlearn, radford2018gpt, peters2018elmo, howard2018ulmfit}, a process in which weights are loaded from a BERT language model that was pre-trained on billions of English tokens. Denote the number of transformer encoder layers as $L$, the hidden size as $H$, and the number of self-attention heads as $A$. The version of BERT used is $\mathrm{{BERT}_{Base}}$ ($L = 12$, $A = 12$, $H = 768$), which has approximately 110-million parameters. Pretrained model weights for BERT can be found on Google Research's BERT Repository.~\footnote{\label{bert-url}\url{https://github.com/google-research/bert}}

\subsubsection{Fully-Connected Dense Layer}

The dense layer is tasked with considering the information passed by BERT's hidden outputs and determining a classification. To accomplish this, it is implemented as a fully-connected neural network layer that accepts input $\mathbf{h}$ and returns $|\mathbf{k}|$ un-normalized activations in $\mathbf{z} \in \mathbb{R}^{|\mathbf{k}|}$, where $\mathbf{k} =  \{0, 1\}$, $\mathbf{z}$ is passed through the softmax normalization function to produce final output vector $\hat{\mathbf{y}}$ as:
\begin{equation} \label{eq:softmax}
    \hat{\mathbf{y}} = \frac{e^{z_i}}{\sum_{j \in \mathbf{k}} e^{z_j}}, \forall i \in \mathbf{k}
\end{equation}

\noindent where each output activation in $\hat{\mathbf{y}}$ represents a classification class. $\hat{\mathbf{y}}$ will later be used to compute the check-worthiness score $CWS$ (Equation \ref{eq:checkworthy}) and compute the predicted classification label as $\hat{y} = \mathrm{argmax}\,\hat{\mathbf{y}}$.

\subsection{Standard Optimization Objective Function} \label{subsec:regularLossFunction}

In neural networks, an objective function, also known as the cost or loss function, is a differentiable expression that serves two purposes: to 1) quantify the disparity between the predicted and ground-truth probability distributions and 2) provide a function for gradient descent to minimize. Negative log-likelihood is a highly common choice for the cost function, because it has a nicely computable derivative for optimization via backpropagation, among other advantageous properties \cite{Janocha_2017}. Our standard negative log-likelihood loss function is formulated as the probability that the model predicts ground-truth $y$ given embedded inputs $\mathbf{s}$, parameterized by the model's weights $\boldsymbol{\theta}$:

\begin{equation} \label{eq:nllCostFunction}
    \mathcal{L}_{reg} = -\frac{1}{N} \sum_{n=1}^{N} \log p(y^{(n)} \mid \mathbf{s}^{(n)}; \boldsymbol{\theta})
\end{equation}

\noindent where $N$ is the total number of training examples in a dataset. $\mathcal{L}_{reg}$ is used to compute adversarial perturbations in Section \ref{subsec:perturbCompute}.

\subsection{Computing Adversarial Perturbations} \label{subsec:perturbCompute}

Gradient-based adversarial training is a regularization technique first introduced in \cite{goodfellow2014explaining}. The procedure trains classifiers to be resistant to small perturbations to its inputs. Rather than passing regular embedded input $\mathbf{s}$ into a processing module such as a transformer or LSTM, adversarial training passes $\mathbf{s}^\prime = \mathbf{s} + \mathbf{r}_{adv}$. $\mathbf{r}_{adv}$ is typically a norm-constrained vector that modifies the input slightly to force the classifier to output incorrect predictions. Then, the disparity between the ground-truth ($y$) and perturbed prediction ($\hat{y}^\prime$) based on the perturbed input is minimized through backpropagation, hence training the model to be resistant to these adversarial perturbations. We are particularly interested in adversarial training's potential as a regularization technique~\cite{shafahi2019adversarial, dalvi2004adversarial, Nguyen_2015, shaham2018understanding, goodfellow2014explaining, miyato2016adversarial}, as BERT networks are prone to overfitting when being fine-tuned on small  datasets~\cite{sun2019fine}. To the best of our knowledge, we contribute the first implementation of this technique on transformer networks.

We denote $\boldsymbol{\theta}$ as the parameterization of our neural network and $\mathbf{r}_{adv}$ as a vector perturbation that is added element-wise to $\mathbf{s}$ before being passed to the transformer encoder. $\mathbf{r}_{adv}$ can be computed in several ways. Firstly, random noise may be added to disrupt the classifier. This is typically formalized as sampling $\mathbf{r}_{adv}$ from a Gaussian distribution: $\mathbf{r}_{adv} \sim \mathcal{N}(\mu, \sigma^2)$.
Alternatively, we can compute perturbations that are \textit{adversarial}, meaning that they increase the model's negative-log-likelihood error (Equation \ref{eq:nllCostFunction}) by the theoretical maximum margin.  This achieves the desired effect of generating a perturbation in the direction in which the model is most sensitive. In this case, $\mathbf{r}_{adv}$ is given as: 
\begin{equation} \label{eq:radvExact}
    \mathbf{r}_{adv} = \operatornamewithlimits{argmax}\limits_{\mathbf{r}, \left\Vert\mathbf{r}\right\Vert \leq \epsilon} \: {-\log p (y \mid \mathbf{s} + \mathbf{r}; \boldsymbol{\theta})}
\end{equation}

\noindent where $\epsilon$ is a constraint on the perturbation that limits the magnitude of the perturbation. 

In \cite{miyato2016adversarial}, it was shown that random noise is a far weaker regularizer than adversarially-computed perturbations. Therefore, we adopt adversarial perturbations for our model (Equation \ref{eq:radvExact}) and propose to apply them on the embeddings of the BERT model. 

Equation \ref{eq:radvExact} gives the \textit{absolute} worst-case adversarial perturbation $\mathbf{r}_{adv}$ given a constraint that $\Vert \mathbf{r} \Vert \leq \epsilon$. However, this value is impossible to compute with a closed-form analytic solution in neural networks; functions such as Equation \ref{eq:nllCostFunction} are neither convex nor concave in topology. Therefore, we propose a novel technique for generating \textit{approximately} worst-case perturbations to the model.

Because BERT embeddings are composed of multiple components (Section \ref{subsec:bert}), it may not be optimal from a regularization standpoint to compute perturbations w.r.t. $\mathbf{s}$. Therefore, to determine the optimal perturbation setting, we propose to experiment with computing $\mathbf{r}_{adv}$ w.r.t. \textit{all} possible combinations of the 3 embedding components. There are 7 different possible configurations in the set of perturbable combinations $\mathcal{P}$, letting $\mathcal{S}$ denote the set of embedding layers: 
\begin{equation} \label{eq:perturbable-set}
    \mathcal{P} = 2^\mathcal{S} - \emptyset \text{ where } \mathcal{S} = \{\mathbf{s}_{tok}, \mathbf{s}_{seg}, \mathbf{s}_{pos}\}
\end{equation}

\begin{figure}[!ht]
  \centering
  \includegraphics[keepaspectratio, width=0.80\columnwidth]{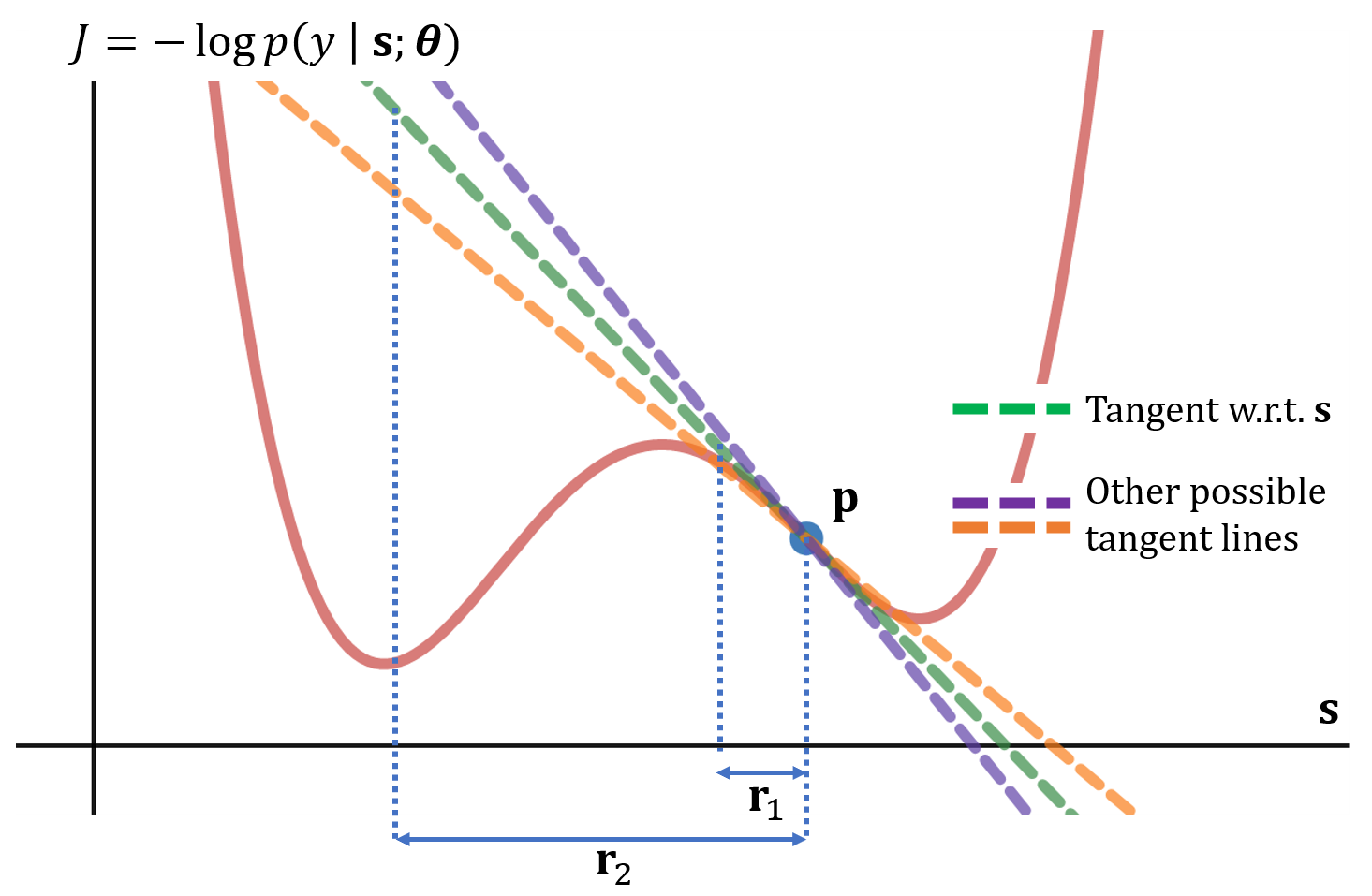}
  \caption{Visualization of Linearization}
  \label{fig:linearization}
\end{figure}

Given this list of components that can be perturbed, we denote sum of the subset of the embeddings we will perturb as $\mathbf{m} = \sum_{x \in \mathbf{b}} x$ where $\mathbf{b} \in \mathcal{P}$. We then generate \textit{approximate} worst-case perturbations by linearizing $\log p(y \mid \mathbf{s}; \boldsymbol{\theta})$ with respect to $\mathbf{m}$. To understand what this means, consider the simplified example shown in Figure \ref{fig:linearization}, which graphs an example cost function $J = -\log p(y \mid \mathbf{s}; \boldsymbol{\theta})$ with respect to an example embedding space $\mathbf{s}$.  For ease of visualization, in Figure \ref{fig:linearization} it is assumed that $\mathbf{s}$ exists on a scalar embedding space; but in reality, our embeddings are in high-dimensional vector space.  The gradient at the point $\mathbf{p}$ gives us information regarding which direction $\mathbf{s}$ should be moved to increase the value of $J$:

\begin{equation} \label{eq:example-perturb-direction}
    \Delta \mathbf{s} \propto \frac{\partial}{\partial \mathbf{m}} \log p(y \mid \mathbf{s}; \boldsymbol{\theta})
\end{equation}

However, we must be careful in determining how much $\mathbf{s}$ should be perturbed, because the assumption that $J$ is linear may not hold in reality. If the perturbation is too large, as with $\mathbf{r}_2$, the adversarial effect will not be achieved, as the value of $J$ will in fact decrease. However, if we introduce a norm constraint $\epsilon$ to limit the perturbations to a reasonable size, linearization can accomplish the task of approximating a worst-case perturbation, as shown with $\mathbf{r}_1$. 

Given the above insight, we generalize the one-dimensional example (Equation \ref{eq:example-perturb-direction}) to higher dimensions using the gradient vector. Therefore, the adversarial perturbation $\mathbf{r}_{adv}$ is computed with Equation \ref{eq:radvApproximate}, which can be implemented using backpropagation in deep neural networks: 
\begin{equation} \label{eq:radvApproximate}
    \mathbf{r}_{adv} = -\epsilon \boldsymbol{\omega} / \left\Vert \boldsymbol{\omega} \right\Vert_2 \text{ where } \boldsymbol{\omega} = \nabla_{\mathbf{m}} -\log p(y \mid \mathbf{s}; \boldsymbol{\theta})
\end{equation}

Since we desire to train our language classification model to become resistant to the perturbations defined in Equation \ref{eq:radvApproximate}, we create adversarially-perturbed input embeddings $\mathbf{s}^\prime$ as follows: 
\begin{equation} \label{eq:sprime}
    \mathbf{s}^\prime = \mathbf{s} + \mathbf{r}_{adv} = \mathbf{s}_{tok} + \mathbf{s}_{seg} + \mathbf{s}_{pos} + \mathbf{r}_{adv}
\end{equation}

After $\mathbf{s}^\prime$ is passed into the transformer module, predictions will be generated. These predictions will be used in formulating the adversarial optimization objective function (Section~\ref{subsec:trainPerturb}). 

\subsection{Compound Optimization Objective} \label{subsec:trainPerturb}
Our model's final optimization objective contains two components: standard loss and adversarial loss. \textit{Standard loss} was defined in Equation \ref{eq:nllCostFunction}. \textit{Adversarial loss} optimizes for distributional smoothness, given by the negative log-likelihood of a model parameterized by $\boldsymbol{\theta}$  predicting $y$ given perturbed input $\mathbf{s}^\prime$: 
\begin{equation} \label{eq:nllCostFunctionAdv}
    \mathcal{L}_{adv} = -\frac{1}{N} \sum_{n=1}^{N} \log p(y^{(n)} \mid \mathbf{s}^{(n)^\prime}; \boldsymbol{\theta})
\end{equation}
\noindent where $N$ represents the number of training samples in $\mathcal{D}$.

The final optimization objective is given as the sum of $\mathcal{L}_{reg}$ and $\mathcal{L}_{adv}$. By combining the two losses, gradient descent will optimize for both distributional smoothness and model accuracy jointly: 
\begin{equation} \label{eq:compoundObjective}
    \operatornamewithlimits{min}\limits_{\boldsymbol{\theta}} \bigg\{
        \mathcal{L}_{reg} + \lambda \mathcal{L}_{adv}
    \bigg\}
\end{equation} 

\noindent where $\lambda$ is a balancing factor between standard and adversarial loss.

\subsection{Adversarial Training Algorithm} \label{subsec:trainproc}

Let $\boldsymbol{\theta}_{tok}$ be the parameters of the token embedding lookup table, $\boldsymbol{\theta}_{seg}$ be the parameters of the segment embedding layer, and $\boldsymbol{\theta}_{pos}$ be the parameters of the positional embedding layer, $\boldsymbol{\theta}_{n}, \forall n < L$ be the parameters for each of the $L$ transformer encoder layers, $\boldsymbol{\theta}_{pool}$ be the parameterization of the pooling layer, and $\boldsymbol{\theta}_{fc}$ be the weights and biases in the fully-connected dense layer. We also define $F$ as the number of encoder layers to freeze (i.e. render the weights uneditable during backpropagation to preserve knowledge obtained from pre-trained weights), where $0 \leq F \leq L$.

\begin{algorithm}
\SetAlgoLined

\SetKwInOut{AlgoInput}{Input}
    \AlgoInput{Training data $\mathcal{D}$}
    \medskip
    Initialize $\boldsymbol{\theta}_{fc}$ using Xavier method~\cite{glorot-xavier-init}\;
    Load pretrained weights for $\boldsymbol{\theta}_{tok}$, $\boldsymbol{\theta}_{seg}$, $\boldsymbol{\theta}_{pos}$, $\boldsymbol{\theta}_n, \forall n < L$\;
    Set $\boldsymbol{\theta}_{tok}$, $\boldsymbol{\theta}_{seg}$, $\boldsymbol{\theta}_{pos}$, and $\boldsymbol{\theta}_{n}, \forall n < F$ to untrainable\;
    $\mathcal{M} \leftarrow$ claim-spotting model (Figure \ref{fig:BERTadv})\;
    \medskip
    \While{not converge}{
        Sample $\mathbf{w}$, $y$ from data $\mathcal{D}$\;
        Tokenize and process $\mathbf{w}$ into $\mathbf{x}$\;
        Pass $\mathbf{x}$ through embeddings to produce $\mathbf{s}$\;
        \medskip
        
        $\triangleright$ Standard forward-propagation\\
        $\hat{y} \leftarrow \mathcal{M}(\mathbf{s};\, \boldsymbol{\theta})$\;
        Compute $\mathcal{L}_{reg}$ using $y, \hat{y}$ (Equation \ref{eq:nllCostFunction})\;
        \medskip
        
        $\triangleright$ Generate and apply perturbations\\
        Compute $\mathbf{r}_{adv}$ using $\mathcal{L}_{reg}, y$ (Equation \ref{eq:radvApproximate})\;
        Compute perturbed input as $\mathbf{s}^\prime = \mathbf{s} + \mathbf{r}_{adv}$\;
        \medskip
        
        $\triangleright$ Adversarial forward-propagation\\
        $\hat{y}^\prime \leftarrow \mathcal{M}(\mathbf{s}^\prime;\, \boldsymbol{\theta})$\;
        Compute $\mathcal{L}_{adv}$ using $y, \hat{y}^\prime$ (Equation \ref{eq:nllCostFunctionAdv})\;
        \medskip
        
        $\triangleright$ Adversarial training\\
        Optimize $\left\{ \mathcal{L}_{reg} + \lambda\mathcal{L}_{adv} \right\}$ (Equation \ref{eq:compoundObjective})\;
    }
    \caption{Adversarial Training Loop}
    \label{algo:adv-train-loop}
\end{algorithm}

The adversarial training procedure is shown in Algorithm \ref{algo:adv-train-loop}. First, model $\mathcal{M}$ is used to compute the optimization function $\mathcal{L}_{reg}$. Then, $\mathcal{L}_{reg}$ is used to compute the adversarial perturbation (Equation~\ref{eq:radvApproximate}), which is used to compute the \textit{adversarial} optimization objective (Equation~\ref{eq:nllCostFunctionAdv}). This objective is added to the standard objective (Equation~\ref{eq:compoundObjective}) and minimized using gradient descent.

%% file: sections/5-results-discussion.tex
\section{Results and Discussions} \label{sec:results-and-discussions}
\begin{table*}[!htb]
\centering
\caption{Sample Sentences, Labels, and Check-Worthiness Scores}
\label{tab:example-sent-label-score-table}
\begin{tabular}{@{}lccc@{}}
\toprule
\multicolumn{1}{c}{\textbf{Claim}}                      & \textbf{Label} & \textbf{CB-SVM CWS Score} & \multicolumn{1}{l}{\textbf{CB-BBA CWS Score}} \\ \midrule
The U.S. loses millions of lives each year to homicide. & CFS            & 0.6000                       & 0.9999                                           \\
I really think you're overthinking the situation.       & NCS            & 0.2178                       & \num{5.8515e-05}                                           \\ \bottomrule
\end{tabular}
\end{table*}
We evaluate our new transformer-based claim-spotting models on both the Classification and Ranking Tasks (Section \ref{subsubsec:taskdef}). We compare against re-trained and refined versions of past ClaimBuster models~\cite{Hassan2017KDDclaimspot,jimenez2018claimspot} and the top-two performing systems from the 2019 CLEF-CheckThat! Challenge. Table~\ref{tab:example-sent-label-score-table} shows several example sentences, their ground-truth labels, and our models' $CWS$ scores.

\subsection{Experiment Setup} \label{subsec:experiment-setup}

\subsubsection{Datasets}

We use two claim-spotting datasets to evaluate model performance.

\medskip \noindent \textbf{ClaimBuster Dataset} (CBD)\label{claimbuster-dataset}: The ClaimBuster dataset is our own in-house, manually labeled dataset. A different version of the CBD was used by~\cite{Hassan2017KDDclaimspot, Hassan2017claimbusterfull, jimenez2018claimspot}. The current CBD consists of two classes, as mentioned in Section \ref{subsubsec:taskdef}: NCS and CFS. The switch to this scheme was motivated by our observation that the non-check-worthy factual sentence class in the previous versions of CBD was not really useful and possibly negatively impacting models trained using it. The CBD consists of 9674 sentences (6910 NCS and 2764 CFS). For validation we perform 4-fold cross validation using this same dataset. The dataset is composed of manually labeled sentences from all U.S. presidential debates from 1960 to 2016. We describe the details of dataset collection in Section~\ref{subsection:dataset-labeling-details}. This dataset is publicly available, as noted in Section~\ref{subsec:datasets}.

\medskip \noindent \textbf{CLEF2019-CheckThat! Dataset} (C$_{2019}$): We also evaluate our model on the 2019 CLEF-CheckThat!~\footnote{\label{clef2019-url}\url{https://sites.google.com/view/clef2019-checkthat/}} claim-spotting dataset.  CLEF-CheckThat! is an annual competition that assesses the state-of-the-art in automated computational fact-checking by providing datasets for claim-spotting. The C$_{2019}$ dataset is comprised of political debate and interview transcripts. Sentences are labelled as check-worthy \textit{only} if they were fact-checked by FactCheck.org. Note that this labelling strategy introduces significant bias into the dataset, as many problematic claims go unchecked due to the limited resources of fact-checkers from a single organization (Section \ref{sec:intro}). The training set contains 15,981 non-check-worthy and 440 check-worthy sentences, and the testing set contains 6,943 non-check-worthy and 136 check-worthy sentences. The C$_{2019}$ dataset also includes speaker information for each sentence, which we did not use in training our models for two reasons: (1) it may introduce unwanted bias based on the name (or guid if names are masked) of speaker and (2) it makes the claim spotting model inapplicable to real-time events since live transcripts typically lack speaker information and speakers not present in the training set are likely to be encountered as well.

\subsection{Evaluated Models} \label{subsec:compared-methods}
\medskip \noindent \textbf{CB-BBA}: This model is trained using our novel claim-spotting framework detailed in Section~\ref{sec:architecture}. It is trained adversarially using the compound optimization objective defined in Equation \ref{eq:compoundObjective}.

\medskip \noindent \textbf{CB-BB}: This model is architecturally identical to CB-BBA but is trained using the standard optimization objective (Equation~\ref{eq:nllCostFunction}). In implementation, $\mathbf{r}_{adv}$ is simply set to $0$. This model serves as a point of comparison for the adversarial model.

\medskip \noindent \textbf{CB-BiL}:~\cite{jimenez2018claimspot} This model is a reimplementation of~\cite{jimenez2018claimspot} in TensorFlow 2.1. It uses normalized GloVe word embeddings~\footnote{\label{glove-url} \url{https://nlp.stanford.edu/projects/glove/}} and consists of a bi-directional LSTM layer which allows it to capture forward and reverse sequence relationships. The model's binary cross entropy loss function is optimized using RMSProp.

\medskip \noindent \textbf{CB-SVM}:~\cite{hassan2015detecting,Hassan2017KDDclaimspot}\label{cb-svm-description}
The SVM classifier uses a linear kernel. The feature-vector used to represent each sentence is composed of a tf-idf weighted bag-of-unigrams vector, part-of-speech vector, and sentence length (i.e., number of words). The total number of features for each sentence using our dataset is $6980$. The core SVM model is produced using scikit-learn's LinearSVC class with the max number of iterations set to an arbitrary high number ($10000000$), to ensure model convergence.

\subsubsection{2019 CLEF-CheckThat! Models}

Neither of the top two teams in CLEF2019 released their code; therefore, we are only able to retrieve their results on CLEF2019.

\medskip \noindent \textbf{Copenhagen}~\cite{Copenhagen-team}: Team Copenhagen's model, the top performer on C$_{2019}$, consisted of an LSTM model~\cite{hochreiter1997long} token embeddings fused with syntactic-dependency embeddings. To train their model, Copenhagen \textit{did not} use the C$_{2019}$ dataset, instead using an external dataset of Clinton/Trump debates that was weakly labeled using our ClaimBuster API.~\cref{claimbuster-url}

\medskip \noindent \textbf{TheEarthIsFlat}~\cite{TheEarthIsFlat-team}: TheEarthIsFlat, the second-place performer, used a feed-forward neural network trained on the C$_{2019}$ dataset. They encoded sentences using the Universal Sentence Encoder~\cite{universal-sentence-encoder}.

\subsection{Embedding Perturbation Study Results} \label{subsec:results-perturbations}
\begin{table}[!htb]
\centering
\captionsetup{justification=centering}
\caption{Perturbation Combinations Study Results\\Averaged Across Stratified 4-Fold Cross Validation}
\label{tab:perturbation-study-table}
\begin{tabular}{@{}ccccccc@{}}
\toprule
                                                        & \multicolumn{2}{c}{\textbf{P}}                                                                & \multicolumn{2}{c}{\textbf{R}}                                                        & \multicolumn{2}{c}{\textbf{F1}}                                                                                    \\ \cmidrule(l){2-7} 
\multirow{-2}{*}{\textbf{ID}}                           & \textbf{NCS}                                      & \textbf{CFS}                                      & \textbf{NCS}                                      & \textbf{CFS}                                      & \textbf{NCS}                                      & \textbf{CFS}                           \\ \midrule
\textbf{0}                                              & 0.9225                                            & 0.8585                                            & 0.9472                                            & 0.8010                                            & 0.9347                                            & 0.8287                                 \\
\textbf{1}                                              & 0.9227                                            & 0.8453                                            & 0.9412                                            & 0.8028                                            & {\color[HTML]{680100} \textbf{0.9319}}            & {\color[HTML]{680100} \textbf{0.8235}} \\
\textbf{2}                                              & 0.9330                                            & {\color[HTML]{680100} \textbf{0.8382}}            & {\color[HTML]{680100} \textbf{0.9357}}            & 0.8321                                            & 0.9344                                            & 0.8351                                 \\
\textbf{3}                                              & 0.9295                                            & 0.8424                                            & 0.9385                                            & 0.8220                                            & 0.9340                                            & 0.8321                                 \\
\textbf{4}                                              & {\color[HTML]{680100} \textbf{0.9201}}            & {\color[HTML]{036400} \textbf{0.8641}}            & {\color[HTML]{036400} \textbf{0.9501}}            & {\color[HTML]{680100} \textbf{0.7938}}            & 0.9349                                            & 0.8275                                 \\
\textbf{5}                                              & 0.9282                                            & 0.8547                                            & 0.9444                                            & 0.8173                                            & {\color[HTML]{036400} \textbf{0.9362}}            & 0.8356                                 \\
\textbf{6}                                              & {\color[HTML]{036400} \textbf{0.9335}}            & 0.8445                                            & 0.9386                                            & {\color[HTML]{036400} \textbf{0.8329}}            & 0.9361                                            & {\color[HTML]{036400} \textbf{0.8386}} \\ \midrule
\textbf{0}                                              & \textbf{1}                                        & \textbf{2}                                        & \textbf{3}                                        & \textbf{4}                                        & \textbf{5}                                        & \textbf{6}                             \\
\begin{tabular}[c]{@{}c@{}}pos\\ seg\\ tok\end{tabular} & \begin{tabular}[c]{@{}c@{}}pos\\ seg\end{tabular} & \begin{tabular}[c]{@{}c@{}}pos\\ tok\end{tabular} & \begin{tabular}[c]{@{}c@{}}seg\\ tok\end{tabular} & pos                                               & seg                                               & tok                                    \\
\bottomrule
\end{tabular}
\end{table}
In Table \ref{tab:perturbation-study-table}, we see the results of perturbing the 3 different embedding layers in BERT. From the results we conclude that setting $6$ produces the best models for our task. Particularly, this setting produces the best recall for the CFS class, which is arguably the more important class. The sacrifice in recall, with respect to the NCS class, compared to other settings is only $\approx0.02$ at most. While setting $4$ achieves the best performance in precision with respect to the CFS class, the drop in recall for the CFS class makes it not viable. Thus, from here on, any results dealing with adversarial training will employ setting $6$ and perturb only the \textit{tok} embedding layer.

\subsection{Classification Task, Ranking, and Distribution Results} \label{subsec:results-binary-ranking}
\begin{table*}[!htb]
\centering
\caption{Precision, Recall, and F1 Averaged Across Stratified 4-Fold Cross Validation}
\label{tab:cb-model-comparison-table}
\begin{tabular}{@{}lcccccccccccc@{}}
\toprule
\multicolumn{1}{c}{\multirow{2}{*}{\textbf{Model}}} & \multicolumn{2}{c}{\textbf{P}} & \multirow{2}{*}{\textbf{P$_{m}$}} & \multirow{2}{*}{\textbf{P$_{w}$}} & \multicolumn{2}{c}{\textbf{R}} & \multirow{2}{*}{\textbf{R$_{m}$}} & \multirow{2}{*}{\textbf{R$_{w}$}} & \multicolumn{2}{c}{\textbf{F1}} & \multirow{2}{*}{\textbf{F1$_{m}$}} & \multirow{2}{*}{\textbf{F1$_{w}$}} \\ \cmidrule(lr){2-3} \cmidrule(lr){6-7} \cmidrule(lr){10-11}
\multicolumn{1}{c}{}                                     & \textbf{NCS}   & \textbf{CFS}  &                                  &                                     & \textbf{NCS}   & \textbf{CFS}  &                                  &                                     & \textbf{NCS}   & \textbf{CFS}   &                                   &                                      \\ \midrule
CB-SVM                                                   & {\color[HTML]{680100} \textbf{0.8935}}             & 0.7972            & 0.8454                               & {\color[HTML]{680100} \textbf{0.8660}}                                  & 0.9263             & {\color[HTML]{680100} \textbf{0.7240}}            & {\color[HTML]{680100} \textbf{0.8251}}                               & {\color[HTML]{680100} \textbf{0.8685}}                                  & 0.9096             & {\color[HTML]{680100} \textbf{0.7588}}             & {\color[HTML]{680100} \textbf{0.8342}}                                & {\color[HTML]{680100} \textbf{0.8665}}                                   \\
CB-BiL                                                & 0.9067             & {\color[HTML]{680100} \textbf{0.7773}}            & {\color[HTML]{680100} \textbf{0.8420}}                               & 0.8697                                  & {\color[HTML]{680100} \textbf{0.9123}}             & 0.7652            & 0.8387                               & 0.8703                                  & {\color[HTML]{680100} \textbf{0.9095}}             & 0.7712             & 0.8403                                & 0.8700                                   \\
CB-BB & {\color[HTML]{036400} \textbf{0.9344}} & 0.8149 & 0.8747 & 0.9003 & 0.9239 & {\color[HTML]{036400} \textbf{0.8379}} & 0.8809 & 0.8993 & 0.9291 & 0.8263 & 0.8777 & 0.8997 \\
CB-BBA & 0.9335 & {\color[HTML]{036400} \textbf{0.8445}} & {\color[HTML]{036400} \textbf{0.8890}} & {\color[HTML]{036400} \textbf{0.9081}} & {\color[HTML]{036400} \textbf{0.9386}} & 0.8329 & {\color[HTML]{036400} \textbf{0.8857}} & {\color[HTML]{036400} \textbf{0.9084}} &
{\color[HTML]{036400} \textbf{0.9361}} & {\color[HTML]{036400} \textbf{0.8386}} & {\color[HTML]{036400} \textbf{0.8873}} & {\color[HTML]{036400} \textbf{0.9082}} \\
\bottomrule
\end{tabular}
\end{table*}

\begin{table}[!htb]
\centering
\captionsetup{justification=centering}
\caption{nDCG Averaged Across\\Stratified 4-Fold Cross Validation}
\label{tab:cb-ndcg-table}
\begin{tabular}{@{}lllll@{}}
\toprule
              & \textbf{CB-SVM}                        & \textbf{CB-BiL} & \textbf{CB-BB}                         & \textbf{CB-BBA} \\ \midrule
\textbf{nDCG} & {\color[HTML]{680100} \textbf{0.9765}} & 0.9817          & 0.9877 & {\color[HTML]{036400} \textbf{0.9894}}          \\ \bottomrule
\end{tabular}
\end{table}

\subsubsection{Classification Results} \label{subsubsec:classification-results}
Our results are encapsulated in Table \ref{tab:cb-model-comparison-table} and Table \ref{tab:cb-ndcg-table}. We assume familiarity with the metrics, which are defined in Section~\ref{subsec:performance-measures}. In Table \ref{tab:cb-model-comparison-table}, we observe that the SVM based model, CB-SVM, has the lowest performance across many measures. This is expected, as the SVM can only capture the information present in the dataset, while the deep learning models benefit from outside knowledge afforded to them by either pre-trained word-embeddings or a pre-trained model (i.e., BERT) that can be fine tuned. The CB-BiL model shows modest improvements overall, but it does achieve noticeably better CFS recall than the SVM model. With respect to BERT-based architectures, both models outperform CB-SVM and CB-BiL considerably. Between CB-BB and CB-BBA, CB-BBA achieves the best performance across most of the metrics. Although CB-BB has a better CFS recall and NFS precision, these differences are minuscule and not enough for CB-BB to be considered a better model. Ultimately, CB-BBA achieves a \textbf{4.70 point F1 score improvement over the past state-of-the-art} CB-BiL model, a 5.31 point F1-score improvement over the CB-SVM model, and a 0.96 point F1-score improvement over a regularly-trained BERT model. This demonstrates the effectiveness of our new architecture and training algorithm.

The results on the C$_{2019}$ dataset are in Table \ref{tab:clef-table}. The metrics presented for the CLEF competition teams are taken from~\cite{clef-checkthat-T1:2019}, since we could not find the source code to reproduce them. For this reason we also cannot provide the P, R, F1, and nDCG for these teams. We tested models trained on both the CBD and C$_{2019}$ training set and used the C$_{2019}$ testing set to evaluate them. The models trained on CBD and tested on the CLEF test set didn't perform as well; this was expected, given that our methodology of dataset labelling differs significantly from CLEF's. Despite this, when trained on C$_{2019}$, CB-BBA held its own when compared to the Copenhagen model which was the winner of that competition only falling short in the mAP metric by 0.0035.

\subsubsection{nDCG Results} \label{subsubsec:ndcg-results}
In Table \ref{tab:cb-ndcg-table} we observe that the best nDCG score is achieved by the CB-BBA model, and the CB-BB and CB-BiL models are within $\approx\frac{6}{1000}$ of it. The CB-SVM model has the ``worst'' nDCG, but is still not far behind the deep learning models. It is noteworthy that all models show relatively good performance on this measure since the CFS class is less represented in the dataset.

\subsubsection{Distribution of $CWS$ Scores} \label{subsubsec:cw-distribution}

\begin{figure}[!htb]
  \centering
  \includegraphics[keepaspectratio, width=1.00\columnwidth]{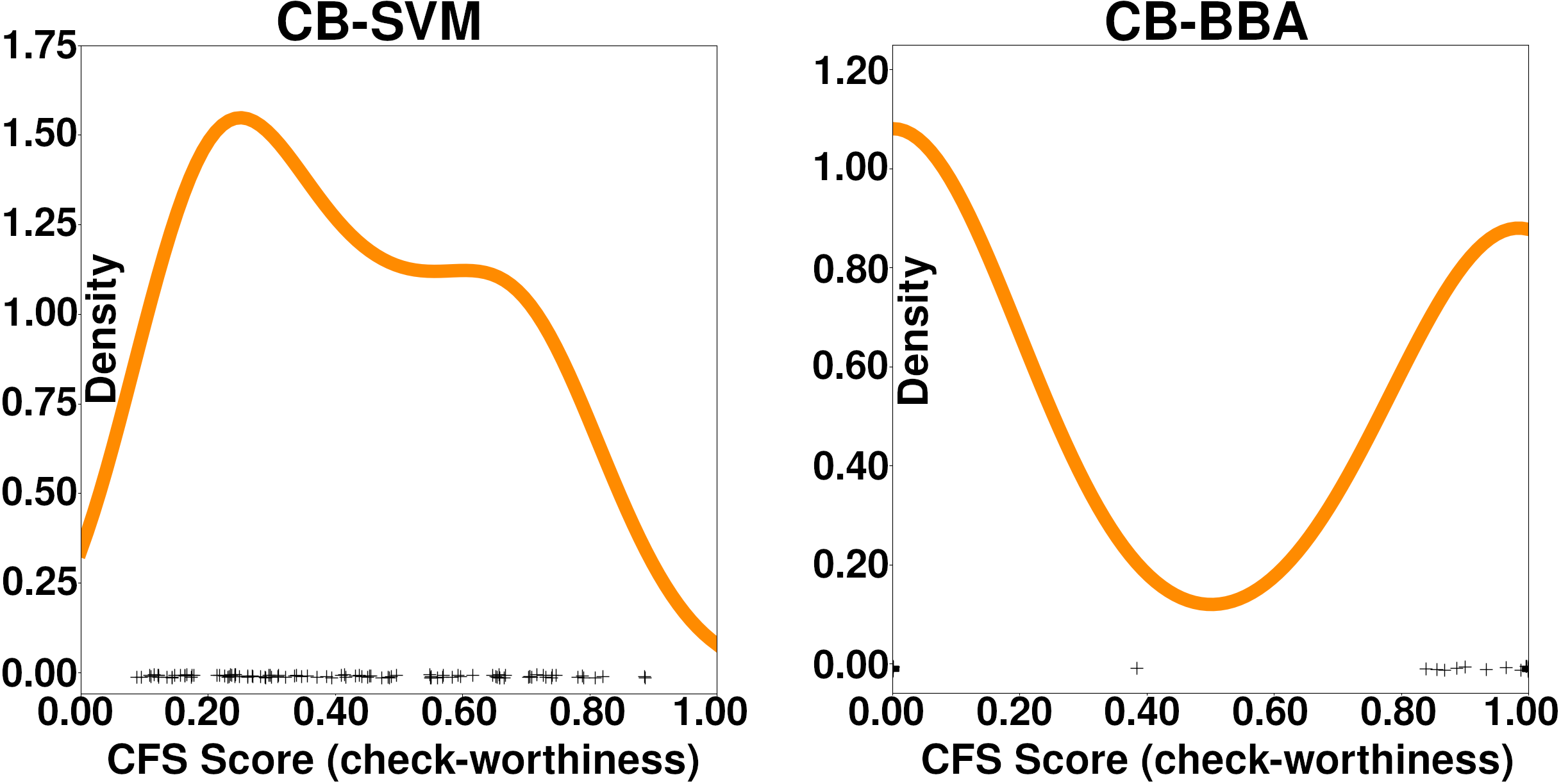}
  \caption{Comparison of Check-Worthiness Score Distributions Using the CB-BBA and CB-SVM Models on 100 sentences from the January 14$\mathrm{^{th}}$, 2020, Democratic presidential debate.}
  \label{fig:cwDistribution}
\end{figure}

To analyze the distribution of our models' outputs on a typical corpus of text, we process 100 sentences from the January 14$\mathrm{^{th}}$, 2020 Democratic presidential debate.~\footnote{\url{https://bit.ly/3bH4fL9}} The sentences were chosen so that there would be about equal numbers of check-worthy and non-check-worthy sentences. Figure \ref{fig:cwDistribution} displays the results, which use Kernel Density Estimation \cite{rosenblatt1956} to estimate the score distribution from discrete data points. Observing the density spikes around 0 and 1 on CB-BBA's distribution, we conclude that our model more clearly differentiates sentences as check-worthy or not check-worthy. The more well-delineated distribution of the CB-BBA model also improves its interpretability over CB-SVM.

\begin{table*}[!htb]
\centering
\caption{CLEF-2019 Test Dataset Classification and Ranking Task Results}
\label{tab:clef-table}
\begin{tabular}{lcccccccccccc}
\toprule
\multicolumn{1}{c}{\multirow{2}{*}{\textbf{Model}}} & \multirow{2}{*}{\textbf{\begin{tabular}[c]{@{}c@{}}Training\\ Dataset\end{tabular}}} & \multirow{2}{*}{\textbf{mAP}} & \multirow{2}{*}{\textbf{P@10}} & \multirow{2}{*}{\textbf{P@20}} & \multirow{2}{*}{\textbf{P@50}} & \multicolumn{2}{c}{\textbf{P}} & \multicolumn{2}{c}{\textbf{R}} & \multicolumn{2}{c}{\textbf{F1}} & \multirow{2}{*}{\textbf{nDCG}} \\ \cline{7-12}
\multicolumn{1}{c}{}                                &                                                                                      &                               &                                 &                                 &                                 & \textbf{NCS}   & \textbf{CFS}  & \textbf{NCS}   & \textbf{CFS}  & \textbf{NCS}   & \textbf{CFS}   &                                \\ \hline
Copenhagen                                          & C$_{2019}$                                                                             & {\color[HTML]{036400} \textbf{0.1660}}                            & 0.2286                              & 0.1571                              & 0.1143                              & \xcancel{\xout{\quad\quad}}              & \xcancel{\xout{\quad\quad}}             & \xcancel{\xout{\quad\quad}}              & \xcancel{\xout{\quad\quad}}             & \xcancel{\xout{\quad\quad}}              & \xcancel{\xout{\quad\quad}}              & \xcancel{\xout{\quad\quad}}                              \\
TheEarthIsFlat                                      & C$_{2019}$                                                                             & 0.1597                            & 0.2143                              & 0.1857                              & {\color[HTML]{036400} \textbf{0.1457}}                              & \xcancel{\xout{\quad\quad}}              & \xcancel{\xout{\quad\quad}}             & \xcancel{\xout{\quad\quad}}              & \xcancel{\xout{\quad\quad}}             & \xcancel{\xout{\quad\quad}}              & \xcancel{\xout{\quad\quad}}              & \xcancel{\xout{\quad\quad}}                             \\
CB-SVM                                              & C$_{2019}$                                                                             & {\color[HTML]{680100}\textbf{0.1087}}                            & 0.1429                              & 0.1429                              & 0.1114                              & 0.9813             & 0.2105            & 0.9978             & 0.0294            & 0.9895             & 0.0516             & {\color[HTML]{680100}\textbf{0.4567}}                             \\
CB-BBA                                     & C$_{2019}$                                                                             & 0.1625                            & {\color[HTML]{036400} \textbf{0.2571}}                              & {\color[HTML]{036400} \textbf{0.1929}}                              & {\color[HTML]{680100}\textbf{0.1086}}                              & {\color[HTML]{680100}\textbf{0.9812}}             & {\color[HTML]{036400} \textbf{0.5000}}            & {\color[HTML]{036400}\textbf{0.9996}}             & {\color[HTML]{680100}\textbf{0.0221}}            & {\color[HTML]{036400}\textbf{0.9903}}             & {\color[HTML]{680100}\textbf{0.0423}}             & {\color[HTML]{036400} \textbf{0.5181}}                             \\
CB-SVM                                              & CBD                                                                          & 0.1134                            & 0.1571                              & 0.1429                              & 0.1143                              & 0.9885             & {\color[HTML]{680100}\textbf{0.0678}}            & 0.8694             & 0.4853            & 0.9251             & 0.1190             & 0.4744                             \\
CB-BBA                                     & CBD                                                                          & 0.1363                            & {\color[HTML]{680100} \textbf{0.1286}}                              & {\color[HTML]{680100}\textbf{0.1286}}                              & 0.1229                              & {\color[HTML]{036400} \textbf{0.9926}}             & 0.0752            & {\color[HTML]{680100}\textbf{0.8354}}             & {\color[HTML]{036400} \textbf{0.6838}}            & {\color[HTML]{680100} \textbf{0.9073}}             & {\color[HTML]{036400} \textbf{0.1356}}             & 0.4978                             \\ \bottomrule
\end{tabular}
\end{table*}

%% file: sections/6-related-works.tex
\section{Related Works} \label{sec:relworks}

In recent years, there have been several efforts with respect to building claim spotting models. ClaimBuster~\cite{hassan2015detecting} is the first of several notable claim spotting models. Another team~\cite{gencheva2017context} extended CB-SVM feature set~\ref{cb-svm-description} by including several contextual features such as: position of the target sentence in its segment, speaker name, interaction between opponents, etc. They created a dataset from the 2016 US presidential and vice presidential debates and annotated sentences by taking fact-checking outputs from 9 fact-checking organizations. If a sentence was fact-checked by at least one fact-checking outlet, it was labeled as check-worthy. A follow-up study~\cite{claimrank} built an online system, namely, ClaimRank~\cref{claimrank-url} for prioritizing sentences for fact-checking based on their check-worthiness score. ClaimRank is a re-implementation of the aforementioned study, but it also supports Arabic by employing cross-language English-Arabic embeddings. Another study~\cite{tathya} followed the same data annotation strategy on a larger dataset by including sentences from an additional 15 2016 U.S. primary debates. The authors developed a multi-classifier based model called TATHYA that uses multiple SVMs trained on different clusters of the dataset. The model takes the output from the most confident classifier. The feature set used was comprised of tf-idf weighted bag-of-unigrams, topics of the sentences, part-of-speech tuples, and a count of entities. 

Another effort by Konstantinovskiy et al.~\cite{konstantinovskiy2018towards}, which utilized the expertise of professional fact-checkers, designed an annotation schema and created a benchmark dataset for training a claim spotting model. The authors trained the model using logistic regression on top of dataset's universal sentence representation derived from InferSent~\cite{infersent}. Their model classifies a sentence as either checkable or non-checkable. The authors also disagreed with ClaimBuster's and ClaimRank's idea of a check-worthiness score. They believe the decision of how important a claim is, should be left to the professional fact-checkers. In the recent CLEF2019 competition on check-worthiness detection, the Copenhagen team developed the winning approach~\cite{Copenhagen-team} which leveraged the semantic and syntactic representation of each word in a sentence. They generated domain-specific pretrained word embeddings that helped their system achieve better performance in the competition. They used a large weakly-labeled dataset, whose labels were assigned by ClaimBuster, for training an LSTM model.

%% file: sections/7-conclusion.tex
\section{Conclusion} \label{sec:conclusion}
We have presented our work on detecting check-worthy factual claims employing adversarial training on transformer networks. Our results have shown that through our methods we have achieved state-of-the-art results on two different datasets (i.e., CBD and C$_{2019}$). During the process we also re-vamped our dataset and approach to collecting and assigning labels for it. We have also come to realize that the lack of a large standardized dataset holds this field back, and thus we look forward to contributing and establishing efforts to fix this situation. We plan on releasing different versions of our dataset periodically in hopes that we can get more significant community contributions with respect to expanding it.~\footnote{\url{https://idir.uta.edu/classifyfact_survey/}}

In the future, we are interested in exploring adversarial training as a \textit{defense against malicious adversaries}. As a publicly deployed API, ClaimBuster may be susceptible to exploitation without incorporating mechanisms that improve its robustness. For example, it has been shown by~\cite{jin2019bert} that a model's classification can be strongly influenced when certain words are replaced by their synonyms. We are currently researching methods to combat similar weaknesses.

%% file: sections/9-reproducibility.tex
\clearpage
\section{Reproducibility} \label{sec:repro}

\subsection{Code Repositories, API, and Related Projects} \label{sec:api-demo-code}

We provide an API of the claim-spotting algorithm online to show its real-world usage at \url{https://idir.uta.edu/claimbuster/api/docs/}. We are also releasing our code along with detailed instructions on running and training our models. Our code and its documentation can be found at \url{https://github.com/idirlab/claimspotter}. Along with these, we also present several projects which showcase how ClaimBuster can and is currently being used:
\begin{itemize}
    \item \href{https://idir.uta.edu/claimbuster/debates}{Claimspotting Presidential Debates~\ExternalLink}
    \item \href{https://idir.uta.edu/claimbuster-dev/factchecker/}{End-to-End Fact-Checking~\ExternalLink}
    \item \href{https://idir.uta.edu/claimportal/}{Claimspotting Tweets~\ExternalLink}
\end{itemize}

\subsection{Formulas for Performance Measures} \label{subsec:performance-measures}
\noindent \textbf{Precision (P):}
\begin{equation} \label{eq:precision}
    \mathrm{P} = \frac{TP}{TP + FP}
\end{equation}

\noindent \textbf{Recall (R):}
\begin{equation} \label{eq:recall}
    \mathrm{R} = \frac{TP}{TP + FN}
\end{equation}

\noindent \textbf{F1:}
\begin{equation} \label{eq:f1}
    \mathrm{F1} = 2 \times \frac{P \times R}{P + R}
\end{equation}

\noindent \textbf{$\mathbf{\left \{ P, R,F1 \right \}}$ Macro (P$_m$, R$_m$, F1$_m$):}
\begin{equation} \label{eq:prf1-macro}
\begin{gathered}
    \mathrm{M}_{m} = \frac{1}{|\mathbf{L}|} \sum_{l \in \mathbf{L}} M_{l}\\
    \textrm{where } M \in \left \{ P, R, F1 \right \} \textrm{, and } L \in \left \{ NCS, CFS \right \}
\end{gathered}
\end{equation}

\noindent \textbf{$\mathbf{\left \{ P, R,F1 \right \}}$ Weighted (P$_w$, R$_w$, F1$_w$):}
\begin{equation} \label{eq:prf1-weighted}
\begin{gathered}
    \mathrm{M}_{w} = \frac{1}{{\sum_{l \in \mathbf{L}} N_l}} \sum_{l \in \mathbf{L}} N_l \times M_{l}\\
    \textrm{where } M \in \left \{ P, R, F1 \right \} \textrm{, } L \in \left \{ NCS, CFS \right \} \textrm{, and}\\
    N_l \textrm{ is the number of samples whose ground truth label is $l$.}
\end{gathered}
\end{equation}

\noindent \textbf{Mean Average Precision (MAP):}
\begin{equation} \label{eq:map}
\begin{gathered}
AP = \frac{\sum_{k=1}^{n} \left ( P\left ( k \right ) \times rel\left ( k \right ) \right ) }{\textit{number of check-worthy claims}}\\
\textrm{where } P\left ( k \right ) \textrm{ is the precision at k, and } rel\left ( k \right ) \textrm{ equals 1 if the}\\
\textrm{claim is check-worthy and 0 otherwise.}\\
MAP = \frac{\sum_{q}^{Q} AP\left ( q \right )}{Q}\\
\textrm{where } Q \textrm{ is the number of queries.}
\end{gathered}
\end{equation}

\noindent \textbf{Normalized Discounted Cumulative Gain (nDCG):}
\begin{equation} \label{eq:ndcg}
\begin{gathered}
nDCG_{p}=\frac{\sum_{i=1}^{p}\frac{2^{rel_{i}}-1}{log_{2}\left ( i+1 \right )}}{\sum_{i=1}^{\left | REL_{p} \right |}\frac{2^{rel_{i}}-1}{log_{2}\left ( i+1 \right )}}\\
\textrm{where } rel_{i}\in \left \{ 0,1 \right \} \textrm{is the \textit{CWS} at position } i \textrm{, and }\\ \left | REL_{p} \right | \textrm{represents the list of claims ordered by their}\\
\textrm{check-worthiness up to position } p\textrm{.}
\end{gathered}
\end{equation}

\subsection{Datasets} \label{subsec:datasets}
We provide the CBD dataset, a collection of sentences labelled manually in-house by high-quality coders, in our repository at \url{https://github.com/idirlab/claimspotter/tree/master/data}. CBD was curated to have $N_{NCS} = Y \times N_{CFS}$, for $Y = 2.5$; where $N_{NCS}$ is the number of non-check-worthy sentences and $N_{CFS}$ is the number of check-worthy sentences. This was done after evaluating different values of $Y$ (i.e., $Y \in \{2, 2.5, 3\}$) and concluding the best ratio for NCS to CFS was $2.5:1$. The C$_{2019}$ dataset, containing sentences from the first and second presidential debates and the first vice presidential debate from 2016, can be found at \url{https://github.com/clef2018-factchecking/clef2018-factchecking}.

\subsubsection{Contributing}
We are always looking for collaborators to contribute to the labelling of more data. Contributions will benefit everyone as we plan on releasing periodic updates when a significant amount of new labels are gathered. To contribute please visit and make an account at:~\url{https://idir.uta.edu/classifyfact_survey/}.

\subsubsection{Dataset Labeling Criteria}\label{subsection:dataset-labeling-details}
The labels for the dataset are assigned by high-quality coders, which are participants that have a pay-rate $\geq 5\textrm{\cent}$ and have labeled at least $100$ sentences. The pay-rate for a user is internally calculated by taking into account their labeling quality, the average length of sentence a user labels, and how many sentences a user skips. More specifically, we define the quality ($LQ_p$) of a coder ($p$) with respect to the screening sentences they have labeled ($SS(p)$) as:
\vspace{-2mm} $$LQ_p = \frac{ \sum_{s \in SS(p)} \gamma^{lt} }{|SS(p)|}$$\vspace{-2mm}\\   
where $\gamma^{lt}$ is the weight factor when $p$ labeled the screening sentence $s$ as $l$ and the experts labeled it as $t$. Both $l,t \in \{NCS$, $CFS\}$. We set $\gamma^{lt} = -0.2$ where $l = t$, $\gamma^{lt} = 2.5$ where $(l,t) \in \{(NCS, CFS), (CFS, NCS)\}$. The weights are set empirically. The pay-rate ($R_p$) is then defined as:
\vspace{-2mm} $$R_p = \frac{L_p}{L}^{1.5}\times(3-\frac{7\times LQ_p}{0.2})\times0.6^{\frac{|SKIP_p|}{|ANS_p|}}$$\vspace{-2mm}\\  
where, $L$ is the average length of all the sentences, $L_p$ is the average length of sentences labeled by $p$, $ANS_p$ is the set of sentences labeled by $p$ and $SKIP_p$ is the set of sentences skipped by $p$. The numerical values in the above equation were set in such a way that it would be possible for a participant to earn up to $10\textrm{\cent}$ per sentence. Using this scheme, out of $581$ users in our system, $69$ users are considered high-quality coders. A label is then only assigned to a particular sentence if it has unanimously been assigned that label by at least 2 high-quality coders. More precisely, we defined the number of high-quality labels needed as: $X \in \left [ NCS, CFS \right ], \exists X \ni s_{X} \geq  2 \wedge s_{X}  = s_{NCS} + s_{CFS}$ where, $s_{X}$ is the number of top-quality labels of type $X$, and a top quality label is one that has been given by a high-quality coder~\cite{Hassan2017KDDclaimspot}.

\subsection{Hyperparameters} \label{subsec:hyperparams}
\begin{table}[!htb]
\caption{Major Parameters for Training}
\label{tab:reproducibility-parameters}
\begin{tabular}{|l|c|c|}
\hline
Parameter                   & \multicolumn{1}{l|}{BBA} & \multicolumn{1}{l|}{BB}  \\ \hline
cs\_train\_steps            & 10                        & 5                       \\ \hline
cs\_lr                      & 5e-5                      & 5e-5                    \\ \hline
cs\_kp\_cls                 & 0.7                       & 0.7                     \\ \hline
cs\_batch\_size\_reg        & 24                        & 24                      \\ \hline
cs\_batch\_size\_adv        & 12                        & -                       \\ \hline
cs\_perturb\_norm\_length   & 2.0                       & -                       \\ \hline
cs\_lambda                  & 0.1                       & -                       \\ \hline
cs\_combine\_reg\_adv\_loss & True                      & -                       \\ \hline
cs\_perturb\_id             & 5                         & -                       \\ \hline
\end{tabular}
\end{table}

We provide Table~\ref{tab:reproducibility-parameters} for major parameter settings used in the BBA, BB, and BiLSTM claimspotting algorithm. The description of the major parameters are as follows: 

\begin{itemize}
    \item cs\_train\_steps: number of epochs to run
    
    \item cs\_lr: learning rate during optimiation
    
    \item cs\_perturb\_norm\_length: norm length of adversarial perturbation
    
    \item cs\_kp\_cls: keep probability of dropout in fully connected layer
    
    \item cs\_lambda: adversarial loss coefficient (eq. \ref{eq:compoundObjective})
    
    \item cs\_combine\_reg\_adv\_loss: add loss of regular and adversarial loss during training
    
    \item cs\_batch\_size\_reg: size of the batch
    
    \item cs\_batch\_size\_adv: size of the batch when adversarial training
    
    \item cs\_perturb\_id: index in Table \ref{tab:perturbation-study-table}
\end{itemize}

\subsection{Evaluation and Training Final Models} \label{subsec:eval-train-final}
We perform 4-fold cross validation to evaluate our models, selecting the best model from each fold using the weighted F1-score (eq. ~\ref{eq:prf1-weighted}) calculated on the validation set. Therefore, in each iteration the data is split as follows: $25\%$ test, $7.5\%$ validation, and $67.5\%$ training. The metrics produced at the end are based on the classifications across all folds. We train the final models (for both CBD and C$_{2019}$) on the entire dataset for up to 10 epochs and select the best epoch based on the weighted F1-score calculated on the validation set.

\subsection{Hardware and Software Specifications}
Our neural network models and training algorithms were written in TensorFlow 2.1~\cite{tensorflowGoogleAI} and run on machines with four Nvidia GeForce GTX 1080Ti GPU's. We did not parallelize GPU usage with distributed training; each experiment was run on a single 1080Ti GPU. The machines ran Arch Linux and had an 8-Core i7 5960X CPU, 128GB RAM, 4TB HDD, and 256GB SSD.